\documentclass[11pt]{article}

\usepackage[final]{acl}

\usepackage{times}
\usepackage{latexsym}

\usepackage[T1]{fontenc}
\usepackage[utf8]{inputenc}

\usepackage{microtype}

\usepackage{inconsolata}

\usepackage{graphicx}

\title{Structure-Preserving Document Translation via Multi-Stage LLM Pipeline: A Case Study in Marathi}

\author{
Manasi Waghe$^{1,2}$ \and
Danish Chandargi$^{1,2}$ \and
Mohammad Aamir Rayyan$^{1,2}$
\AND
Raviraj Joshi$^{2,3}$ \and
A.R. Deshpande$^{1}$ \\
$^{1}$Pune Institute of Computer Technology, Pune, India \\
$^{2}$L3Cube Labs, Pune, India \\
$^{3}$Indian Institute of Technology Madras, Chennai, India \\
\texttt{manasi.waghe12@gmail.com, danishchandargi@gmail.com} \\
\texttt{aamirray19@gmail.com, ravirajoshi@gmail.com, ardeshpande@pict.edu}
}

\begin{document}
\maketitle
\begin{abstract}
Government documents in India are predominantly issued in regional languages such as Marathi, creating substantial accessibility barriers for non-native readers, interstate administrative bodies, and policy analysts. Although recent advances in neural machine translation have improved sentence-level translation quality, existing systems largely neglect document structure, formatting integrity, and domain-specific terminology, thereby limiting their applicability to official documentation. This paper presents a structure-preserving Marathi-to-English government document translation framework capable of performing end-to-end document transformation while maintaining layout fidelity. The proposed system integrates layout-aware optical character recognition, coordinate-based text extraction, large language model based translation, and structured document reconstruction through HTML representations. By enforcing spatial alignment constraints and preserving hierarchical document elements, the framework ensures structural consistency between the source and translated documents. Experimental evaluation on real-world Marathi government PDFs demonstrates improved structural preservation, translation coherence, and terminological consistency compared to conventional text-only translation pipelines. The proposed framework contributes toward scalable multilingual accessibility solutions for e-governance and administrative document processing.
\end{abstract}

\section{Introduction}

Government agencies across India routinely publish administrative orders, policy notifications, circulars, and regulatory documents in regional languages such as Marathi \cite{joshi2022l3cube_mahacorpus}. While these documents serve as authoritative sources of governance, linguistic barriers significantly restrict their accessibility for non-native speakers, interdepartmental coordination, and broader public engagement. Manual translation processes remain resource-intensive and slow, while existing automated translation systems typically operate at the plain-text level without preserving document structure.

A critical limitation of current machine translation pipelines is their inability to maintain structural components such as tabular layouts, hierarchical headings, enumerated clauses, paragraph alignment, and official formatting conventions. Consequently, translated outputs often lose their visual coherence and administrative usability. This problem is further compounded by the presence of domain-specific legal and bureaucratic terminology, complex syntactic constructions, and rigid stylistic conventions commonly observed in government documentation.

Recent large language models (LLMs) such as GPT-based systems have demonstrated strong multilingual translation capabilities due to their contextual understanding and reasoning abilities \cite{gala_indictrans2:_2023,fan_beyond_2020,team_no_2022}. However, these models typically operate on unstructured text and lack inherent awareness of document layout, spatial constraints, and formatting hierarchy. When applied directly to document translation, LLMs may produce structurally inconsistent outputs by altering paragraph organization, numbering patterns, or formatting relationships. Such limitations make purely LLM-based approaches insufficient for official government documents where structural fidelity is essential.

Another technical challenge arises from language expansion effects during translation. Marathi text translated into English frequently results in increased token length, causing spatial misalignment when directly substituted into the original document layout. Most translation systems lack coordinate-aware reconstruction mechanisms, resulting in structural distortions and formatting inconsistencies.

To address these challenges, this work proposes a layout-preserving Marathi-to-English document translation framework specifically designed for government PDFs. Unlike conventional approaches that treat translation as an isolated linguistic task, the proposed framework models document translation as a structure-aware transformation problem. The system integrates layout-aware OCR, neural translation, coordinate-constrained text alignment, and structured document reconstruction to produce translated outputs that retain both semantic fidelity and visual consistency.

The principal contributions of this work include:

\begin{enumerate}
\item Development of a layout-aware document translation pipeline integrating OCR, translation, and reconstruction.
\item A coordinate-constrained text reinsertion mechanism for structural preservation.
\item Domain-robust translation using large language models.
\item A parallelized processing architecture for improved computational efficiency.
\item A deployable framework for large-scale government document translation.
\end{enumerate}

By jointly optimizing linguistic accuracy and document structure preservation, the proposed framework enables the generation of translated documents suitable for official administrative workflows and multilingual governance systems.

\section{Literature Survey}

Recent developments in neural machine translation have substantially improved translation quality for Indic languages. Models such as IndicTrans2 demonstrate strong multilingual capabilities across scheduled Indian languages including Marathi-English translation\cite{gala_indictrans2:_2023}. However, these models primarily operate at the sentence level and do not address challenges related to document layout preservation or formatting reconstruction.

Large-scale multilingual translation systems such as M2M-100 and similar approaches extend translation capabilities across hundreds of languages without requiring pivot languages\cite{fan_beyond_2020,team_no_2022}. While these models demonstrate strong linguistic generalization, they are not optimized for domain-specific administrative language and often fail to preserve structural integrity when applied to formatted documents.

Document intelligence models such as LayoutLM and LayoutLMv3 integrate visual and textual embeddings to enable structured document understanding\cite{xu_layoutlm:_2020,xu_layoutlmv2:_2022,huang_layoutlmv3:_2022}. Similarly, architectures like DocFormer further enhance multimodal document representation learning\cite{appalaraju_docformer:_2021}. These models have demonstrated effectiveness in tasks such as form understanding, document classification, and key information extraction. However, their architectures are not designed for document translation or layout-preserving text reinsertion.

Recent advances in OCR systems have enabled structured extraction of text along with positional metadata. Traditional OCR engines such as Tesseract provide foundational text extraction capabilities but lack robust layout understanding. More recent layout-aware approaches improve spatial reasoning but typically do not incorporate translation into a unified document transformation pipeline.

Large language models have also been explored for translation and document post-editing tasks due to their strong contextual reasoning abilities \cite{touvron_llama:_2023,zhao_survey_2026}. While these models achieve high linguistic fluency, they often lack explicit mechanisms for preserving spatial layout and formatting constraints. Unconstrained LLM translation may also introduce paraphrasing or structural inconsistencies, which are undesirable in administrative and legal contexts where textual precision and formatting stability are critical. These observations suggest that LLMs are most effective when integrated into structure-aware document processing pipelines rather than used as standalone document translation solutions.

Research in document transformation workflows has also explored intermediate representations such as HTML and XML for preserving structural relationships during processing \cite{xu_layoutlm:_2020,xu_layoutlmv2:_2022,huang_layoutlmv3:_2022,appalaraju_docformer:_2021}. These approaches demonstrate that maintaining structural metadata significantly improves reconstruction fidelity. However, existing systems rarely integrate OCR, translation, and layout reconstruction into a cohesive end-to-end architecture.

From the existing body of research, it is evident that prior work addresses either translation quality or document structure preservation independently. Limited research has explored integrated systems combining layout-aware OCR, domain-aware translation, and coordinate-guided reconstruction. The proposed work seeks to bridge this gap through a unified framework capable of preserving both semantic correctness and structural fidelity.

\section{Methodology}
\subsection{System Architecture}

The proposed framework follows a modular pipeline architecture designed to perform layout-preserving Marathi-to-English document translation while maintaining both semantic accuracy and structural fidelity \cite{gala_indictrans2:_2023,xu_layoutlm:_2020,xu_layoutlmv2:_2022,huang_layoutlmv3:_2022}. Unlike conventional translation systems that operate purely on extracted text, the proposed approach preserves spatial metadata throughout the processing pipeline to ensure accurate document reconstruction.

\begin{figure}[t]
    \includegraphics[width=\columnwidth]{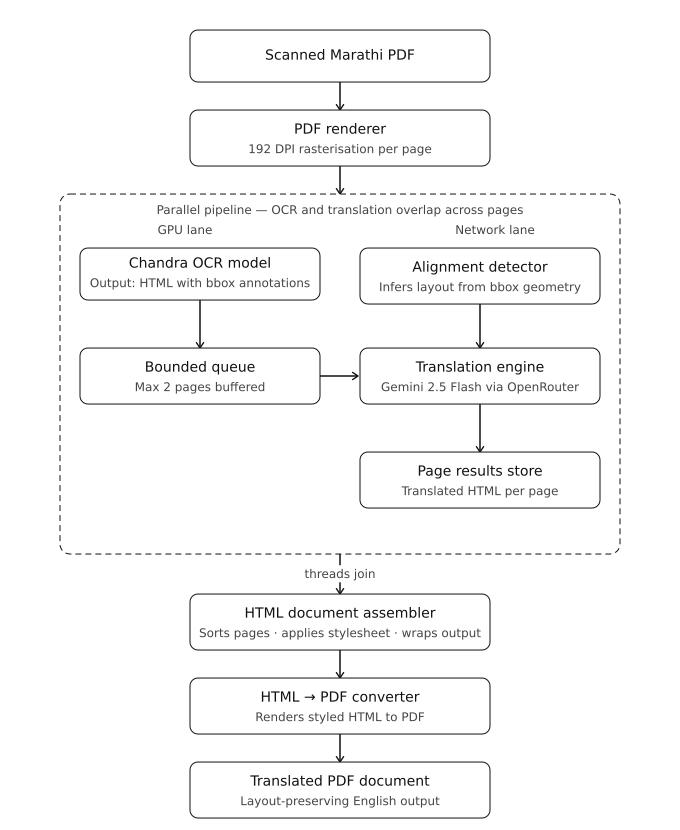}
    \caption{System architecture of the proposed layout-preserving Marathi-to-English document translation framework.}
    \label{fig:system_architecture}
\end{figure}

The overall workflow is illustrated in Fig.~\ref{fig:system_architecture}. The pipeline begins with PDF rasterization followed by layout-aware OCR extraction. The structured OCR output is then processed through parallel OCR and translation stages before final document reconstruction. The architecture emphasizes three key principles: structure preservation through bounding box metadata, pipeline parallelism for efficiency, and modular design for extensibility.

Following OCR extraction, structured text blocks are translated while preserving layout constraints, and the translated content is then reassembled into the final document.

\subsection{Document Ingestion and Preprocessing}

The translation pipeline begins by processing the input Marathi PDF using PyMuPDF to extract individual pages. Each page is rendered as a high-resolution image to improve OCR accuracy and ensure reliable detection of text regions. This approach allows the system to handle both digitally generated and scanned documents \cite{xu_layoutlm:_2020,xu_layoutlmv2:_2022,huang_layoutlmv3:_2022}.

The document can be represented as:

\begin{equation}
D = \{P_1, P_2, \dots , P_n\}
\end{equation}

where each page is processed independently before final aggregation. This page-level processing improves scalability and enables parallel execution for large documents.
\subsection{Layout-Aware Text Extraction}

The system uses Chandra OCR\footnote{Available online at Hugging Face: \href{https://huggingface.co/datalab-to/chandra}{datalab-to/chandra}} to extract both textual content and layout information \cite{smith2007tesseract}. Unlike traditional OCR systems that return plain text, this model produces structured HTML containing positional metadata and document hierarchy \cite{xu_layoutlm:_2020,xu_layoutlmv2:_2022,huang_layoutlmv3:_2022,appalaraju_docformer:_2021}.

Each text region is represented as:

\begin{equation}
B = (T, C)
\end{equation}

where T represents text and C represents bounding box coordinates defined as:

\begin{equation}
C = (x_0, y_0, x_1, y_1).
\end{equation}

This representation preserves reading order and spatial relationships between document elements, enabling accurate reconstruction after translation.

\subsection{Neural Translation}

Extracted text blocks are translated using a large language model through an API-based inference framework \cite{touvron_llama:_2023,zhao_survey_2026}. Translation is performed at the block level to preserve contextual continuity and avoid fragmentation errors \cite{zhao_survey_2026,li_augmenting_2020}.

The translation process can be defined as:

\begin{equation}
T_{eng} = MT(T_{mar})
\end{equation}

where T\_mar denotes Marathi text, T\_eng denotes translated English text, and MT denotes the translation model.

To maintain structural consistency, only textual content is translated while layout tags and formatting attributes remain unchanged. This ensures that translation does not disrupt document structure.

\subsection{Spatial Alignment and Reinsertion}

Following translation, the system reinserts translated text into the original document structure using bounding box constraints. This step is necessary because English translations often occupy more space than the source Marathi text.

The alignment process ensures that translated text remains within its original spatial boundaries using adaptive wrapping and overflow control. This process can be represented as:

\begin{equation}
B' = f(T_e, C)
\end{equation}

where B' represents the reconstructed block.

This coordinate-guided reinsertion is critical for maintaining document readability and preventing layout distortion \cite{xu_layoutlm:_2020,xu_layoutlmv2:_2022,huang_layoutlmv3:_2022,appalaraju_docformer:_2021}.

\subsection{Parallel Processing Strategy}

To improve efficiency, the system employs a producer–consumer pipeline in which OCR extraction and translation operate concurrently. OCR processing generates structured text blocks which are placed into a processing queue, while the translation module consumes these blocks asynchronously.

This design reduces idle time between processing stages and improves throughput, particularly for multi-page documents. The bounded queue also ensures balanced resource utilization across GPU and API processing stages.

\subsection{Document Reconstruction}

After translation and alignment, the document is reconstructed using structured HTML representations. Translated pages are assembled while preserving layout attributes and formatting relationships \cite{xu_layoutlm:_2020,xu_layoutlmv2:_2022,huang_layoutlmv3:_2022,appalaraju_docformer:_2021}.

The reconstructed document can be represented as:

\begin{equation}
O = \{R_1, R_2 \dots R_n\}
\end{equation}

where $R_i$ represents each reconstructed translated page.

This reconstruction process ensures preservation of paragraph alignment, structural hierarchy, and formatting consistency with the source document.

\subsection{Post-Processing}

The final stage performs minor refinements including encoding normalization, removal of OCR artifacts, and formatting corrections. These adjustments improve document readability and ensure the translated output meets practical usability requirements.

\subsection{Implementation Details}

The framework is implemented using PyMuPDF for document processing, Chandra OCR for layout extraction, large language model APIs for translation, BeautifulSoup for structured parsing, and Python threading for pipeline parallelism \cite{smith2007tesseract,touvron_llama:_2023,zhao_survey_2026}. The modular design allows easy replacement or improvement of individual components.

\section{Results and Analysis}

\begin{figure}[t]
    \includegraphics[width=\columnwidth]{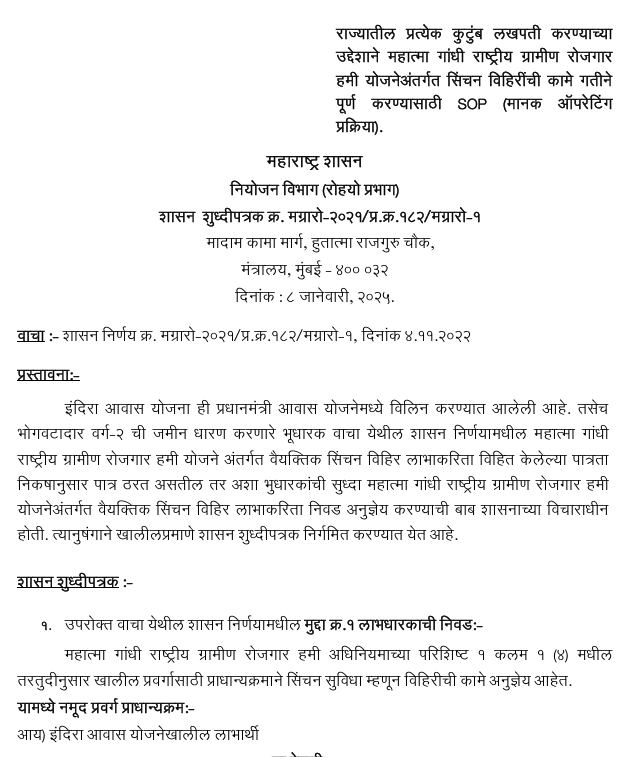}
    \caption{Original Marathi Government document}
    \label{fig:original_doc}
\end{figure}

\begin{figure}[t]
    \includegraphics[width=\columnwidth]{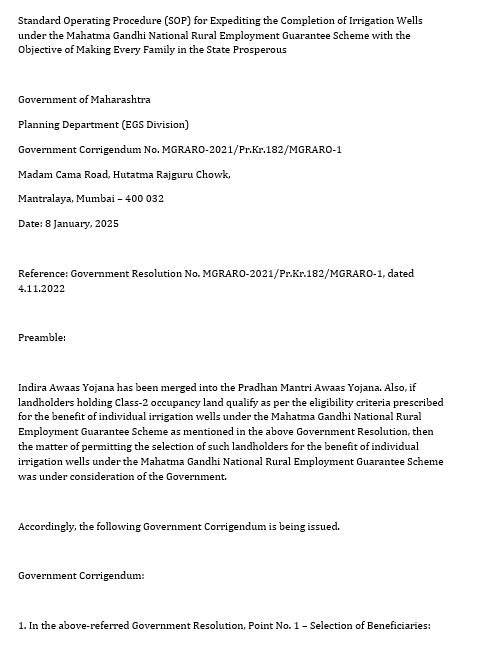}
    \caption{LLM translation without layout constraints showing structural distortion.}
    \label{fig:llm_output}
\end{figure}

\begin{figure}[t]
    \includegraphics[width=\columnwidth]{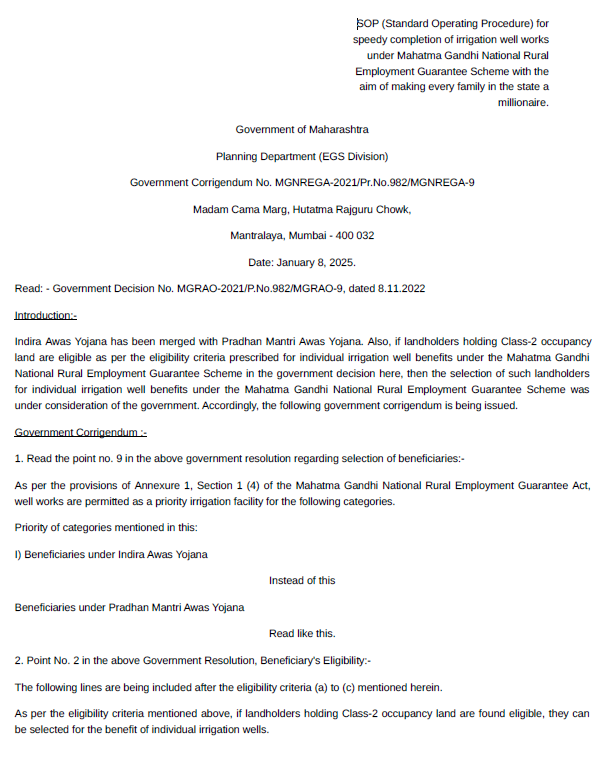}
    \caption{Layout-preserving translation produced by the proposed framework.}
    \label{fig:proposed_output}
\end{figure}

\subsection{Experimental Setup}

The proposed framework was evaluated on a collection of real-world Marathi government documents including administrative circulars, public notices, and policy documents. The dataset contained documents with complex layouts such as hierarchical headings, tables, mixed formatting, and structured paragraphs \cite{zhong_publaynet:_2019,xu_layoutlm:_2020,xu_layoutlmv2:_2022,huang_layoutlmv3:_2022}. Both digitally generated and scanned PDFs were included to evaluate robustness across document types.

The evaluation focused on three aspects: translation quality, layout preservation, and document readability. Since document translation involves both linguistic and structural fidelity, evaluation was primarily qualitative and based on comparison with baseline approaches rather than relying solely on traditional translation metrics.

The system was compared against three common baselines: direct text extraction followed by translation, OCR with naive text replacement, and LLM-based translation without layout constraints. Experiments were conducted in a GPU-enabled environment for OCR, while translation was performed using API-based inference. The evaluation emphasized practical usability of translated documents.

\subsection{Translation Quality Analysis}

Translation quality was assessed based on semantic accuracy, contextual preservation, and consistency of administrative terminology. Government documents often contain formal expressions that require context-aware translation rather than literal sentence conversion.

Block-level translation improved contextual continuity compared to sentence-level approaches by preserving logical groupings of administrative text. The use of large language models also improved grammatical fluency and readability while maintaining consistency in recurring official terminology.

Compared to baseline OCR-translation pipelines, the proposed approach produced more coherent translations and reduced the need for manual corrections. Additionally, comparison with direct LLM-based document translation showed that while LLMs provide strong linguistic translation quality, they do not inherently preserve document structure when formatting constraints are not explicitly enforced.

\subsection{Layout Preservation Evaluation}

A major objective of this work was to preserve document structure during translation. Results indicate that maintaining coordinate metadata throughout the pipeline significantly improves reconstruction quality \cite{xu_layoutlm:_2020,xu_layoutlmv2:_2022,huang_layoutlmv3:_2022,appalaraju_docformer:_2021}.

The translated documents preserved paragraph ordering, heading alignment, and structural hierarchy more effectively than baseline methods, where formatting was often disrupted. The coordinate-aware reinsertion strategy ensured translated text remained within its designated spatial regions.

To further illustrate this difference, a qualitative comparison was performed between direct LLM-based translation and the proposed framework. While the LLM output produced linguistically correct translations, it failed to preserve document layout, resulting in loss of formatting, inconsistent spacing, and broken paragraph structure. In contrast, the proposed system maintained both semantic correctness and structural organization. This comparison is illustrated in Fig.~\ref{fig:original_doc} and Fig.~\ref{fig:llm_output}.

The intermediate HTML representation further contributed to structural preservation by separating content from layout information \cite{xu_layoutlm:_2020,xu_layoutlmv2:_2022,huang_layoutlmv3:_2022,appalaraju_docformer:_2021}. These results highlight the importance of treating layout preservation as a core component of document translation rather than a secondary refinement step.

\subsection{Readability and Document Usability}

The translated documents remained visually organized and readable with minimal manual intervention. The adaptive text placement strategy reduced overflow and alignment issues typically caused by translation-induced text expansion \cite{xu_layoutlm:_2020,xu_layoutlmv2:_2022,huang_layoutlmv3:_2022}.
Although minor spacing adjustments were occasionally required in dense sections, the overall document structure remained stable. Compared to baseline methods, the proposed system significantly reduced manual reformatting effort, improving its suitability for administrative workflows.

A visual comparison between the original document, LLM translation output, and the proposed system output (Fig.~\ref{fig:original_doc}–Fig.~\ref{fig:proposed_output}) further demonstrates that preserving layout information directly improves readability and practical usability of translated documents.

\subsection{Performance Analysis}

The pipeline parallelism in the proposed architecture improved processing efficiency by allowing OCR and translation stages to operate concurrently. This reduced idle processing time and improved throughput for multi-page documents.

The producer–consumer design allowed translation to begin immediately after OCR extraction, reducing sequential bottlenecks. The modular architecture also supports scalability for larger document collections.

\subsection{Limitations}

Some limitations were observed during evaluation. OCR inaccuracies in low-quality scanned documents may affect downstream translation quality \cite{smith2007tesseract,baek_what_2019}. Documents containing dense tables also remain challenging to perfectly reconstruct due to space constraints.

Additionally, large text expansion during translation may occasionally require minor formatting adjustments. Future improvements could address these challenges through glossary enforcement, domain adaptation, and improved layout adjustment techniques.

\section{Conclusion}

This paper presented a layout-aware Marathi-to-English government document translation framework designed to address the limitations of conventional text-only translation systems. By treating document translation as a combined structural and linguistic transformation problem, the proposed system preserves both meaning and formatting.

The integration of layout-aware OCR, neural translation, coordinate-guided alignment, and structured reconstruction enables the generation of translated documents with improved structural fidelity and readability. The results demonstrate that preserving spatial metadata significantly improves document usability compared to traditional translation pipelines.

The study also highlights the importance of structure-aware translation for administrative documents, where formatting is essential for maintaining document integrity. Experimental comparison with direct LLM-based translation further shows that linguistic quality alone is insufficient for document translation tasks that require structural fidelity.

Overall, the framework provides a practical solution for improving multilingual accessibility of government documents while reducing manual translation effort. Future work will focus on improving table reconstruction, incorporating domain-specific terminology control, extending support to additional Indic languages, and exploring multimodal document models for improved structural understanding.

\section*{Acknowledgements}
This work was carried out under the mentorship of L3Cube Labs, Pune. We would like to express our gratitude towards our mentor for his continuous support and encouragement. This work is a part of the L3Cube-MahaNLP project\cite{joshi2022l3cube_mahanlp}.

\nocite{*}
\bibliography{main}
\end{document}